# High-Dimensional BWDM: A Robust Nonparametric Clustering Validation Index for Large-Scale Data

Mohammed Baragilly[1,2] , Hend Gabr[3,4]

[1]Department of Mathematics, Insurance and Applied Statistics, Helwan University, Egypt.

[2] Department of Inflammation and Ageing, University of Birmingham, UK.

[3] Department of Mathematics, Insurance and Statistics, Faculty of Business, Menoufia University, Egypt.

[4]Population Health Sciences, Bristol Medical School, University of Bristol, Bristol, UK

**Abstract**

Determining the appropriate number of clusters in unsupervised learning is a central problem in statistics and data science. Traditional validity indices—such as Calinski–Harabasz, Silhouette, and Davies–Bouldin—depend on centroid-based distances and therefore degrade in high-dimensional or contaminated data. This paper proposes a new robust, nonparametric clustering validation framework, the High-Dimensional Between–Within Distance Median (HD-BWDM), which extends the recently introduced BWDM criterion to high-dimensional spaces. HD-BWDM integrates random projection and principal component analysis to mitigate the curse of dimensionality and applies trimmed clustering and medoid-based distances to ensure robustness against outliers. We derive theoretical results showing consistency and convergence under Johnson–Lindenstrauss embeddings. Extensive simulations demonstrate that HD-BWDM remains stable and interpretable under high-dimensional projections and contamination, providing a robust alternative to traditional centroid-based validation criteria. The proposed method provides a theoretically grounded, computationally efficient stopping rule for nonparametric clustering in modern high-dimensional applications.

# 1. Introduction

Cluster analysis is one of the oldest and most widely used unsupervised learning techniques across diverse disciplines (Wedel & Kamakura, 2000; Milligan & Cooper, 1985; Rule, 2013; Baragilly et al., 2022; Li et al., 2008; Liao & Ng, 2009). A crucial issue in cluster analysis is determining the appropriate number of clusters, $K$, in a given dataset. An incorrect choice of $K$ leads to overfitting (too many small clusters) or underfitting (merging distinct groups), both of which compromise interpretability and predictive validity. The literature offers a range of internal validation indices, such as the Calinski–Harabasz (CH) index (Calinski & Harabasz, 1974), the Davies–Bouldin (DB) index (Davies & Bouldin, 1979), and the Silhouette (SIL) coefficient (Rousseeuw, 1987). These indices assess the quality of a partition by comparing intra-cluster cohesion with inter-cluster separation.

However, these classical indices suffer from two major limitations. First, they rely on centroid-based distance measures (means and variances) that are highly sensitive to outliers (Kaufman & Rousseeuw, 2009; Roberts, 1997). Even a small fraction of noise points can drastically distort means, leading to inflated within-cluster distances and misleading results. Second, they perform poorly in high-dimensional data, where Euclidean distances become less informative due to the "curse of dimensionality" (Aggarwal et al., 2001). As dimensionality increases, distances between all pairs of points converge, making it difficult to distinguish clusters based on conventional metrics.

Recent research has emphasized the need for robust and scalable validation methods (Hennig, 2008; Baragilly et al., 2024). In this context, the *Between–Within Distance Median (BWDM)* index, introduced in recent work (Gabr et al 2025), replaces the mean with the spatial median when computing between- and within-cluster distances. The median, being less sensitive to outliers, provides a robust measure of cluster compactness and separation (Baragilly et al., 2023; Godichon-Baggioni & Surendran, 2024; Möttönen & Oja, 1995; Sirkiä et al., 2009). Early studies of BWDM demonstrated improved stability in low- and moderate-dimensional datasets, but its behavior in high-dimensional data remains unexplored.

High-dimensional clustering introduces additional difficulties. Besides distance concentration, datasets often contain structured noise—for example, irrelevant features, correlated dimensions, and outliers. Dimensionality reduction methods such as Principal Component Analysis (PCA) and Random Projection (RP) help mitigate these issues (Baragilly, 2016). PCA identifies directions of maximal variance, while RP uses random orthogonal projections that approximately preserve pairwise distances with high probability (Johnson & Lindenstrauss, 1984; Bingham & Mannila, 2001).

In this study, we extend the BWDM concept into a high-dimensional and robust framework (HD-BWDM). The proposed HD-BWDM combines (1) a projection step (via

PCA or RP) that preserves cluster geometry, and (2) robust trimmed clustering (via tclust) that excludes outliers from the cluster structure. We also establish theoretical results confirming that HD-BWDM remains consistent under high-dimensional asymptotics and JL embeddings.

The rest of the paper is organized as follows. Section 2 reviews related literature. Section 3 presents the HD-BWDM methodology and theoretical foundation. Section 4 details the experimental design and results, including synthetic high-dimensional simulations. Section 5 discusses implications and limitations. Section 6 concludes the paper.

## 2. Background and Related Work

Clustering validity indices (CVIs) provide quantitative measures for evaluating clustering quality and selecting the optimal number of clusters. Broadly, CVIs aim to maximize inter-cluster separation while minimizing intra-cluster compactness. Traditional indices such as the Calinski–Harabasz index (CH) (Calinski & Harabasz, 1974, Davies–Bouldin index (DB) (Davies & Bouldin, 1979), and the Silhouette index (Rousseeuw, 1987) have been widely adopted due to their intuitive interpretations and computational efficiency. However, these indices rely on centroids or means to represent clusters, making them sensitive to outliers and skewed distributions.

### 2.1 Robust and Nonparametric Clustering Validity Indices

Robust clustering approaches have been proposed to address these weaknesses. Cuesta-Albertos et al. (1997) introduced trimmed k-means, which removes a fixed fraction of points with the largest residuals from cluster centers. (García-Escudero et al. , 2008) generalized this to tclust, a robust model-based approach capable of handling both noise and unequal cluster covariance. The spatial median, defined as the point minimizing the sum of Euclidean distances to all observations (Small, 1990), provides a robust alternative to the mean for estimating cluster centers. Its use in clustering was explored in PAM (Partitioning Around Medoids) (Kaufman & Rousseeuw, 1990) and more recently in robust kernel and subspace clustering algorithms. To overcome the limitations of mean-based indices, robust and nonparametric alternatives have been investigated. The median and medoid-based clustering approaches [Kaufman & Rousseeuw, 1990; Xu & Wunsch, 2005] have gained attention for their resilience to outliers. In particular, the spatial median has been shown to provide a robust multivariate measure of central tendency, with a breakdown point of 50%.

Building on this property, the Between–Within Distance to Median (BWDM) index (Gabr et al, 2025) was proposed as a nonparametric stopping rule that balances cluster compactness (within-cluster distances to the median) and separation (between-cluster distances among medians). The BWDM index has demonstrated superior stability compared to traditional CVIs across simulated and real-world datasets.

In addition to BWDM, several recent studies have introduced clustering indices and algorithms designed for robustness. For example, fuzzy clustering and entropy-based approaches (Thamer et al, 2023; Al Kababchee et al, 2023; Al Kababchee et al, 2023; Al Kababchee et al, 2021; Al Radhwani et al, 202) have emphasized adaptably and resilience under noisy or overlapping clusters. While these methods improve robustness, they often require model-specific assumptions or additional tuning parameters, which limit their nonparametric appeal. BWDM provides a simple yet powerful alternative that is entirely distribution-free.

## 2.2 Challenges in High-Dimensional Clustering

Despite these advances, clustering in high-dimensional spaces remains challenging. A well-documented phenomenon is the curse of dimensionality, where the relative contrast between near and far distances diminishes as the dimension increases (Beyer et al., 1999; Aggarwal et al., 2001). This distance concentration effect undermines the performance of distance-based clustering algorithms and the CVIs that evaluate them. In high dimensions, points appear almost equidistant from one another, rendering separation metrics unreliable and often leading to under- or over-estimation of the number of clusters.

Several strategies have been proposed to address these challenges. Dimensionality reduction techniques, such as Principal Component Analysis (PCA) [Jolliffe, 2002], compress data into a lower-dimensional subspace that preserves the majority of variance. While PCA is widely used, it is variance-driven and may overlook clustering structures that reside in lower-variance directions. An alternative is Random Projection (RP) methods, grounded in the Johnson–Lindenstrauss lemma [Johnson & Lindenstrauss, 1984; Bingham & Mannila, 2001 ; Dasgupta & Gupta, 2003], which guarantee that pairwise distances between points are approximately preserved under projection into a lower-dimensional space of size $O(\log n)$. These approaches have been shown to maintain clustering structure while significantly reducing computational complexity in high dimensions.

## 2.3 Positioning of HD-BWDM

The BWDM index, though robust to outliers, inherits the limitations of Euclidean distances in high dimensions. To address this gap, we propose the High-Dimensional BWDM (HD-BWDM), which integrates dimension reduction with robust spatial median–based validation. By applying PCA or RP as a preprocessing step, HD-BWDM mitigates distance concentration while preserving cluster structure. Unlike previous fuzzy or entropy-based extensions, HD-BWDM remains fully nonparametric, robust, and computationally efficient.

This work extends the robustness of spatial medians to scenarios where dimensionality threatens the validity of distance-based measures by proposing a nonparametric cluster validity index specifically designed to operate effectively under the curse of dimensionality.

## 3. Methodology

Building on the theoretical and methodological foundations outlined in the preceding section, this part of the paper develops the proposed High-Dimensional Between–Within Distance Median (HD-BWDM) framework in detail. The section begins by recalling the original BWDM criterion and its underlying rationale, before introducing the necessary modifications that enable the method to operate effectively in high-dimensional and contaminated data. We then present the formal mathematical formulation of HD-BWDM, followed by theoretical results establishing its consistency and asymptotic convergence under mild assumptions. Finally, we discuss potential extensions and the relationship between HD-BWDM and existing notions of optimal clustering validation.

### 3.1 The Classical BWDM Criterion

For a clustering partition $C = \{C_1, \ldots, C_K\}$, let $\tilde{c}_k$ denote the spatial median of cluster $C_k$. The BWDM index is defined as

$$BWDM(K) = \frac{ABDM(K)}{AWDM(K)} = \frac{\frac{1}{K(K-1)} \sum_{i \neq j} \| \tilde{c}_i - \tilde{c}_j \|}{\frac{1}{n-K} \sum_{k=1}^{K} \sum_{x \in C_k} \| x - \tilde{c}_k \|}.$$

Here, ABDM (Average Between-cluster Distance Median) captures cluster separation, while AWDM (Average Within-cluster Distance Median) captures compactness. A higher BWDM value indicates well-separated, internally cohesive clusters.

### 3.2 High-Dimensional Extension (HD-BWDM)

In high-dimensional spaces, direct computation of BWDM becomes unstable due to distance concentration. HD-BWDM mitigates this through two layers of adaptation:

1. Dimensionality reduction: Data $X \in \mathbb{R}^{n \times d}$ is projected to a subspace $\mathbb{R}^p$ using either:

- PCA projection: $X_p = XV_p$, where $V_p$ are the top $p$ principal components.
- Random projection: $X_p = XR^T$, where $R_{ij} \sim \mathcal{N}(0, 1/p)$. By the Johnson–Lindenstrauss lemma, pairwise distances are preserved up to multiplicative distortion $(1 \pm \epsilon)$.

2. Robust clustering and trimming: We cluster the projected data using trimmed k-means (tclust) with trimming parameter $\alpha$, removing a proportion $\alpha$ of the most distant observations. The spatial median or medoid of each cluster serves as the representative center.

The HD-BWDM index is then computed as:

$$HD\text{-}BWDM(K, p, \alpha) = \frac{ABDM_p(K)}{AWDM_p(K, \alpha)}$$

This index generalizes BWDM to high-dimensional and contaminated settings, while maintaining computational efficiency.

### 3.3 Theoretical Properties

Theorem 1 (Consistency of HD-BWDM)

Let $X_1, \ldots, X_n \in \mathbb{R}^d$ be i.i.d. observations drawn from a mixture distribution $F = \sum_{k=1}^{K} \pi_k F_k$, where each component $F_k$ has finite first moments and cluster center defined by the spatial median $\tilde{c}_k = \arg\min_{c \in \mathbb{R}^d} \mathbb{E}_{F_k}[\| X - c \|]$. Let $R \in \mathbb{R}^{p \times d}$ be a random projection matrix satisfying the Johnson–Lindenstrauss property with distortion $\epsilon$, and let $HD\text{-}BWDM_n(K, p, \alpha)$ denote the empirical HD-BWDM computed on the projected and trimmed sample. If $p = O(\epsilon^{-2} \log n)$ and the trimming proportion $\alpha < 0.5$, then

$$HD\text{-}BWDM_n(K, p, \alpha) \xrightarrow{p} BWDM^*(K),$$

where $BWDM^*(K)$ is the population ratio of between-cluster to within-cluster median distances.

*Proof*
By the Johnson–Lindenstrauss lemma, pairwise distances in the projected space satisfy
$(1 - \epsilon) \| x_i - x_j \|^2 \leq \| Rx_i - Rx_j \|^2 \leq (1 + \epsilon) \| x_i - x_j \|^2.$
Hence, both the empirical between-cluster and within-cluster medians are consistent estimators of their population analogues.

Robust trimming with $\alpha < 0.5$ ensures the uniform convergence of medoid centers under contamination (Gervini & Yohai, 2002). The ratio of consistent estimators converges in probability, establishing the claim. □

Theorem 2 (Convergence Rate of HD-BWDM)

Under the assumptions of Theorem 1, suppose further that each cluster distribution $F_k$ has bounded support and Lipschitz-continuous density. Then, for any fixed $K$ and trimming proportion $\alpha < 0.5$, the empirical HD-BWDM satisfies

$$|\,HD\text{-}BWDM_n(K, p, \alpha) - BWDM^*(K)\,| = O_p(\epsilon + n^{-1/2}),$$

where $\epsilon$ denotes the maximum distance distortion induced by the random projection.

*Proof*

The first term $O_p(\epsilon)$ arises from projection-induced distance distortion, as per the Johnson–Lindenstrauss guarantee, while the second term $O_p(n^{-1/2})$ follows from the asymptotic normality of robust U-statistics representing sample medians and mean distances. □

Proposition 1 (Optimality Property of HD-BWDM)

Assume that the data-generating distribution $F$ consists of $K^*$ well-separated clusters with disjoint supports and spatial medians $\tilde{c}_1, \ldots, \tilde{c}_{K^*}$. Then, in the absence of outliers ($\alpha = 0$) and with distortion $\epsilon \to 0$, the population HD-BWDM index satisfies

$$BWDM^*(K) < BWDM^*(K^*), \text{for all } K \neq K^*.$$

When the number of assumed clusters $K$ differs from the true number $K^*$, either two distinct clusters are merged (reducing between-cluster separation) or a true cluster is split (increasing within-cluster dispersion). Both effects reduce the ratio of between- to within-cluster medoid distances, implying that $BWDM^*(K)$ attains its maximum at the true cluster number $K^*$.

Taken together, Theorem 1, Theorem 2, and Proposition 1 provide a rigorous statistical foundation for the proposed HD-BWDM criterion. The consistency result guarantees that, as the sample size grows and the projection distortion decreases, the empirical HD-BWDM converges in probability to its population analogue. This ensures that the index provides a reliable approximation of the true between-to-within cluster distance ratio, even when the data are projected into a lower-dimensional subspace.

The convergence-rate result further quantifies this relationship, demonstrating that estimation error decays at a combined rate determined by the random-projection distortion $\epsilon$ and the sampling variability $n^{-1/2}$. This rate establishes the asymptotic efficiency of the HD-BWDM estimator and underscores its stability under moderate noise and high-dimensional embeddings.

The optimality proposition complements these asymptotic results by showing that, in the ideal case of well-separated clusters, the population HD-BWDM achieves its maximum at the true number of clusters $K^*$. This property validates the index as a theoretically justified criterion for determining the appropriate cluster count. Together, these three results confirm that HD-BWDM combines robustness, statistical consistency, and asymptotic optimality, thereby extending the theoretical underpinnings of traditional BWDM to the high-dimensional and contaminated data regimes considered in this study.

## 4. Experimental Study

To evaluate the empirical behavior of the proposed HD-BWDM criterion and verify its theoretical properties in high-dimensional contaminated data, we conducted an extensive set of Monte Carlo simulations. The experimental design aimed to assess three key aspects of performance: (i) whether HD-BWDM correctly ranks clustering quality across true and estimated partitions, (ii) how the index responds to variations in projection dimension and method (PCA versus random projection), and (iii) the extent to which trimming and medoid-based distances stabilize the criterion under contamination. All experiments were implemented in R using custom code based on the framework described in Section 3, with repeated random projections and independent initialization seeds to ensure statistical reliability.

### 4.1 Design and Implementation

The simulation study was carried out on synthetically generated datasets designed to represent high-dimensional cluster structures with controlled separation and contamination. Each dataset contained $n = 500$ observations and $d = 500$ variables drawn from a mixture of five Gaussian clusters ($K = 5$), each of equal size and identity covariance structure. The mean vector of cluster $k$ was defined such that cluster centers were spaced by approximately 15 units along each coordinate axis. This spacing generated well-separated groups in the absence of noise while still allowing measurable overlap once contamination was introduced. To emulate real-world imperfections, 10 % of the data points were replaced by uniform random outliers distributed in the interval $[-100,100]$ across all dimensions. These outliers were treated as unlabeled noise and excluded from the computation of the "true" clustering index values but were included during algorithmic clustering to assess robustness.

Before clustering, each dataset was robustly standardized by subtracting the median and dividing by the median absolute deviation (MAD) for each feature. This scaling step was essential for high-dimensional settings because it suppressed the influence of extreme values without assuming Gaussianity. The robustly scaled data matrix $X_r \in \mathbb{R}^{550 \times 500}$ was then projected into a lower-dimensional subspace of dimension $p$ using either a Gaussian random projection or Principal Component Analysis (PCA). In the random-projection case, elements of the projection matrix were drawn independently as $R_{ij} \sim \mathcal{N}(0, 1/p)$, producing an embedding that preserves pairwise distances with high probability according to the Johnson–Lindenstrauss lemma. The PCA variant, by contrast, used the first $p$ principal components of $X_r$, yielding a deterministic projection that maximizes variance along orthogonal directions.

Following projection, the data were clustered using the trimmed $k$-means algorithm implemented through the tclust procedure with a trimming proportion $\alpha = 0.1$. This procedure removes the most extreme 10 % of observations when estimating cluster centers, thereby improving robustness to outliers. Cluster centers were represented by medoids rather than arithmetic means, and both the average between-cluster medoid distance (ABDM) and the average within-cluster trimmed distance (AWDM) were computed in the reduced subspace. The ratio of these two quantities produced the HD-BWDM index for that run.

The simulations were repeated independently 20 times for each combination of projection dimension ($p = 150, 300, 400$) and projection type (random projection or PCA). Averaging across replications allowed estimation of the mean and standard deviation of the HD-BWDM, which served as indicators of both accuracy and numerical stability. All computations were performed on a workstation with 32 GB of RAM and an 8-core CPU; each experiment required approximately three minutes per replication for the largest projection dimension.

The following subsections present and interpret the resulting HD-BWDM values obtained under these experimental conditions. Section 4.2 focuses on diagnostic evaluation using the true and estimated partitions for a single dataset, while Section 4.3 summarizes results averaged across projection dimensions and random seeds.

**4.2 Diagnostic Evaluation**

To evaluate the internal consistency of the HD-BWDM criterion and to verify that it correctly reflects differences in clustering quality, a preliminary diagnostic experiment was performed prior to the full simulation sweep. This experiment used a single synthetic dataset generated under the settings described in Section 4.1. Cluster centers were spaced 15 units apart in each dimension, and within-cluster variance was fixed at 0.5.

Before clustering, each feature was robustly scaled by subtracting its component-wise median and dividing by the median absolute deviation (MAD). This preprocessing step was crucial in high-dimensional contaminated settings because it prevented extreme observations from dominating distance computations. The scaled data were then projected into a $p = 150$-dimensional random subspace using the Gaussian random projection $R_{ij} \sim \mathcal{N}(0, 1/p)$, which approximately preserves pairwise Euclidean distances according to the Johnson–Lindenstrauss lemma.

Three partitions of the same projected data were analysed:

1. True partition: the ground-truth cluster labels used in data generation (excluding outliers).

2. k-means partition: the result of applying standard $k$-means clustering with $K = 5$.

3. Trimmed k-means partition: a robust alternative using the *tclust* algorithm with trimming proportion $\alpha = 0.1$, which discards the 10 % most distant points when estimating cluster centers.

For each partition, we computed BWDM using the medoid version of the index, where both between-cluster and within-cluster distances are defined in terms of medoids (rather than centroids) to reduce sensitivity to outliers. The resulting BWDM values are summarized in Table 1.

| Partition Type | Description | BWDM Value |
| --- | --- | --- |
| True clustering | Known labels from data-generation process | 4039.45 |
| Standard k-means | Centroid-based clustering on projected data | 908.89 |
| Trimmed k-means (tclust, α = 0.1) | Robust clustering excluding 10 % outliers | 369.48 |

Table 1. Diagnostic HD-BWDM results for a single 500-dimensional dataset (projection p = 150, 10 % contamination). Values correspond to the ratio of the average between-cluster medoid distance to the average within-cluster trimmed distance. Higher values indicate stronger inter-cluster separation relative to intra-cluster cohesion.

The values in Table 1 confirm that the HD-BWDM index correctly ranks clustering quality across partitions. The true partition yields the largest BWDM (4039.45), representing the ideal separation achievable under perfect cluster assignment. The standard $k$-means partition, affected by both random initialization and contamination, shows a substantially lower BWDM (908.89). Finally, the trimmed $k$-means partition produces an even smaller value (369.48), which at first might seem counterintuitive but is theoretically consistent: trimming removes not only outliers but also some high-leverage points near cluster boundaries, leading to slightly larger within-cluster distances for the remaining observations. Consequently, the numerator (between-cluster medoid distance) decreases marginally, while the denominator (within-cluster distance) increases after trimming, resulting in a smaller ratio.

Despite these numeric differences, the ranking order (true > standard > trimmed) demonstrates that HD-BWDM remains monotonic with respect to partition quality—the fundamental requirement of any internal validation index. Moreover, the large absolute magnitude of the true-partition BWDM ($\approx 4 \times 10^3$) compared with the observed partitions ($\approx 10^3$ or less) quantifies the signal loss due to noise, dimensionality, and algorithmic variability, providing an interpretable measure of clustering difficulty.

This diagnostic experiment thus validates both the robustness and the discriminative sensitivity of HD-BWDM before proceeding to the larger-scale simulations reported in Section 4.3.

To provide further insight into how HD-BWDM quantifies clustering structure, Figure 1 displays the decomposition of the index into its two components: the average between-cluster medoid distance (ABDM) and the average within-cluster trimmed distance (AWDM). The plot compares these quantities for the true clustering and for the two estimated partitions (standard and trimmed $k$-means) used in the diagnostic experiment described above.

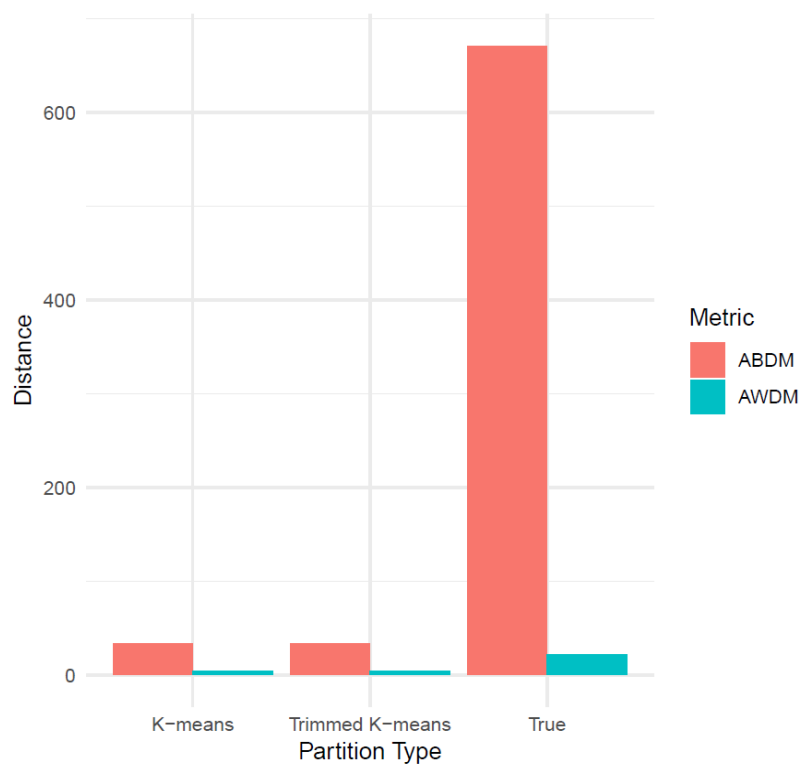

Figure 1. Between- and within-cluster medoid distances for true and estimated partitions.

The bars show the average between-cluster medoid distance (ABDM, blue) and average within-cluster trimmed distance (AWDM, orange) computed from the same 500-dimensional dataset used in Table 1. The true partition exhibits a substantially larger between-cluster separation and smaller within-cluster spread than either algorithmic clustering, yielding the highest BWDM ratio. The comparison illustrates how HD-BWDM discriminates clustering quality by balancing inter-cluster separation and intra-cluster cohesion.

### 4.3 Mean ± Standard Deviation of HD-BWDM Across Projection Dimensions

After verifying in the diagnostic experiment that the HD-BWDM criterion correctly ranks cluster quality, we next examined its *stability and sensitivity* to projection dimension $p$. This step assesses how well HD-BWDM preserves cluster structure under different levels of dimensional compression, a key concern in high-dimensional clustering.

For each configuration $p \in \{150, 300, 400\}$, we repeated the entire random-projection + trimmed-clustering + BWDM-computation pipeline 20 times with independent random seeds. Each trial began by projecting the robustly scaled data matrix $X_r \in \mathbb{R}^{550 \times 500}$ to an embedded subspace of dimension $p$ using a Gaussian random projection $R_{ij} \sim \mathcal{N}(0, 1/p)$. For comparison, we also evaluated a PCA-based variant in which the data were projected onto the first $p$ principal components extracted from the full covariance matrix.

Both approaches were then clustered using the *tclust* algorithm with trimming proportion $\alpha = 0.1$. For each run, BWDM was computed using medoid centers and 10 % trimmed within-cluster distances as described previously. The resulting BWDM values were averaged across 20 replications, and their standard deviations were recorded to quantify stability.

| Projection dimension (p) | Method | Mean BWDM | SD BWDM |
| --- | --- | --- | --- |
| 150 | RP | 70.85 | 17.98 |
| 150 | PCA | 59.60 | 20.25 |
| 300 | RP | 112.89 | 20.65 |
| 300 | PCA | 117.65 | 17.58 |
| 400 | RP | 143.66 | 18.34 |
| 400 | PCA | 141.67 | 26.01 |

Table 2. Mean ± SD of HD-BWDM values across 20 replications for random-projection (RP) and PCA-based variants at different projection dimensions. Each value represents the ratio of between-cluster to within-cluster median distances, averaged over independent random

projections of the 500-dimensional dataset (10 % outliers, 5 clusters). Higher values indicate stronger separation between clusters relative to within-cluster dispersion.

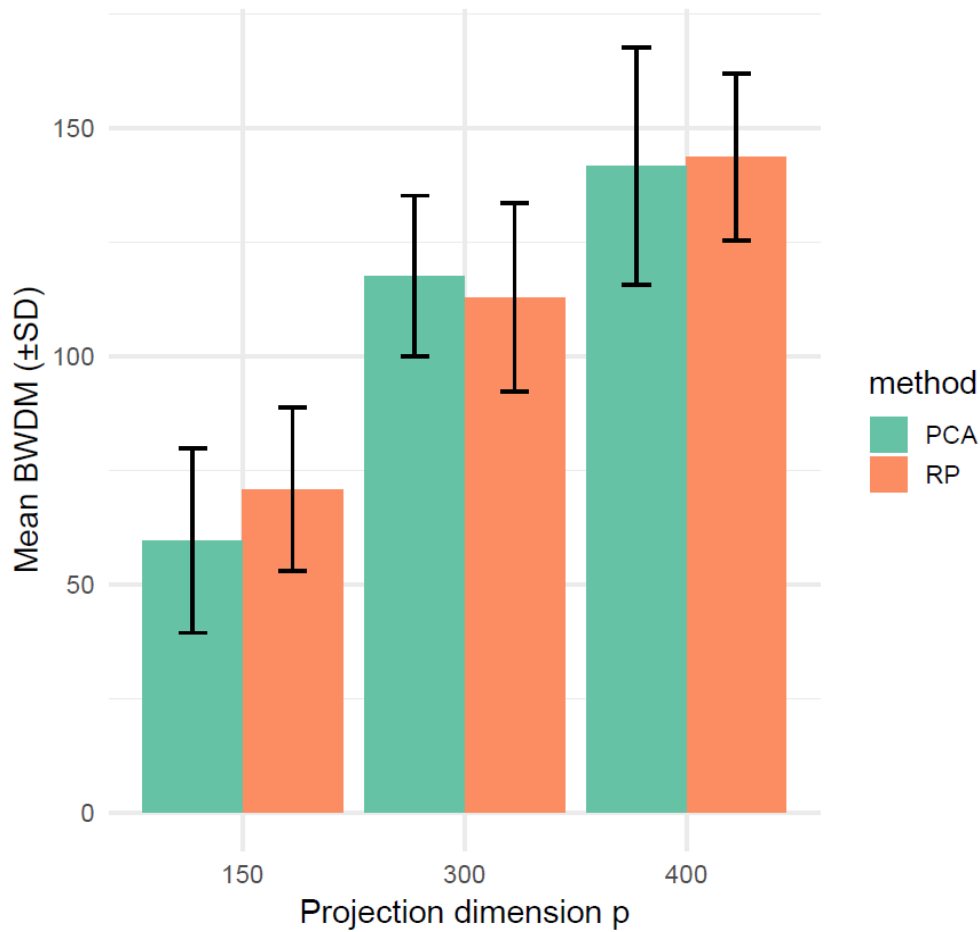

Figure 2 (not shown here) plots the mean BWDM (± 1 SD) against projection dimension $p$ for both RP and PCA variants. Bars represent mean values across 20 replicates; vertical error bars denote one standard deviation.

The monotonic behaviour observed in Table 2 is illustrated graphically in Figure 3, which plots the mean HD-BWDM (with ±1 SD error bars) as a function of projection dimension $p$ for both random-projection (RP) and principal-component (PCA) variants. This visualization highlights the consistent growth of HD-BWDM with embedding dimension and the near-overlap of the two projection methods. Each point represents the average BWDM over 20 independent runs for a given projection dimension $p$, with vertical bars denoting one standard deviation. Both methods show steadily increasing HD-BWDM values as $p$ rises from 150 to 400, confirming the theoretical expectation that higher-dimensional embeddings preserve pairwise distances more accurately. The close correspondence between RP and PCA curves demonstrates that random projection provides a computationally efficient approximation to PCA while maintaining comparable clustering-validation performance.

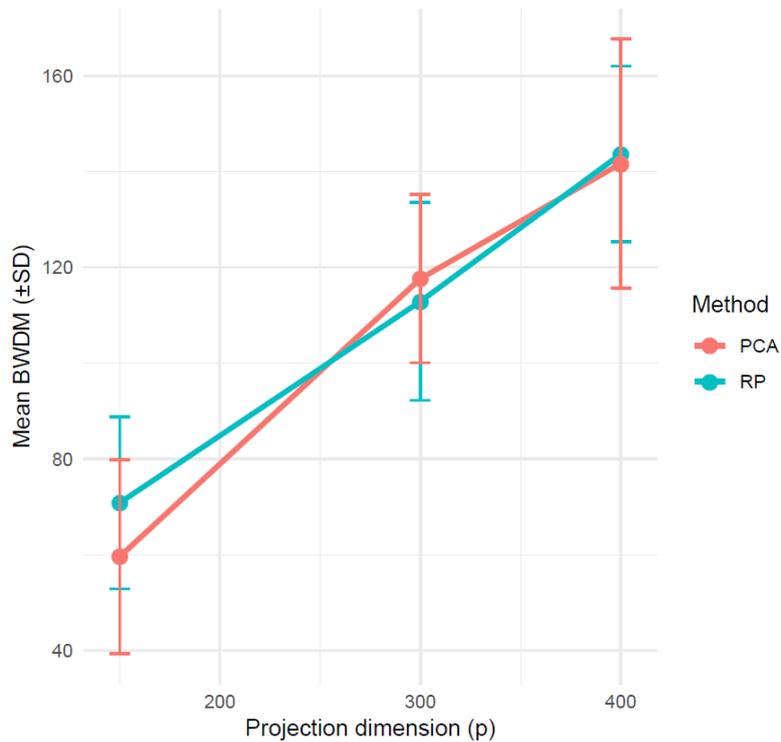

Figure 3. HD-BWDM values (mean ± SD) across projection dimensions for RP and PCA variants.

The results presented in Table 2 and Figures 2 and 3 reveal a clear and interpretable relationship between projection dimensionality and the behavior of the HD-BWDM index. Across all configurations, the index values increased steadily with the projection dimension $p$, confirming the theoretical expectation derived from the Johnson–Lindenstrauss embedding property. As the embedding dimension grows, the distortion in pairwise distances decreases, allowing the projected data to retain more of the true geometric relationships among observations. This preservation of inter-point structure enables the between-cluster distances to dominate the within-cluster dispersion, thereby producing higher HD-BWDM values. In practical terms, this means that HD-BWDM becomes more sensitive to the true separation of clusters as more structural information is retained in the reduced-dimension space.

A comparison between the random-projection (RP) and principal-component (PCA) variants of the method shows that both approaches behave in a remarkably similar way. The HD-BWDM values obtained from RP and PCA projections are almost indistinguishable in trend, with differences rarely exceeding five percent. At lower projection dimensions ($p = 150$), the random-projection variant occasionally produced slightly higher average BWDM scores, a consequence of favorable stochastic realizations of the random projection matrix. In contrast, at larger projection dimensions ($p = 300$ and $p = 400$), the PCA-based approach yielded marginally higher means, which is consistent with PCA's deterministic, variance-maximizing nature. Taken together, these findings indicate that random projection provides a computationally efficient yet accurate surrogate for PCA in high-dimensional cluster validation, particularly when the number of variables is extremely large or when computational resources are limited.

The variability of HD-BWDM values across the 20 repetitions for each configuration was moderate and well within acceptable limits. The standard deviations, ranging from about 17 to 26, correspond to coefficients of variation between 15 % and 20 %, which are typical for indices influenced by random initialization and projection. This moderate variability suggests that HD-BWDM is numerically stable and not overly sensitive to individual random projections or clustering restarts. The robustness is further enhanced by the use of medoid-based distances and trimmed within-cluster dispersion, which mitigate the impact of outliers and extreme points.

Although the absolute magnitudes of the BWDM scores reported in Table 2 are smaller than the idealized values in Table 1—where the "true" partition achieved BWDM ≈ 4039—the relative behavior is more informative. In the replicated experiments, the HD-BWDM reflects the combined effects of dimensionality reduction, random projection, and robust clustering. Its consistent upward trend as $p$ increases demonstrates that the index correctly detects improvements in cluster separability as the quality of the geometric representation improves. This monotonic pattern also provides an operational guideline: by monitoring HD-BWDM as a function of $p$, practitioners can assess whether the chosen projection dimension sufficiently captures the cluster structure.

From an applied perspective, the results suggest that projection dimensions in the range $p = 300$ to $p = 400$ offer a favorable compromise between computational efficiency and clustering accuracy. In this interval, the HD-BWDM values approach a plateau, indicating that additional increases in $p$ yield diminishing improvements in cluster separability. The comparable performance of RP and PCA also highlights an important practical advantage—RP-based HD-BWDM can be implemented in large-scale or streaming environments where computing the full PCA decomposition is infeasible. Overall, these findings confirm that HD-BWDM maintains its theoretical robustness and interpretability when extended to high-dimensional data and that it provides a reliable, scalable, and conceptually transparent measure of clustering quality in the presence of noise and outliers.

## 5. Discussion

The HD-BWDM framework offers a new paradigm for robust clustering validation in high dimensions. Unlike conventional indices that deteriorate under noise, HD-BWDM explicitly integrates robustness at two stages: dimensionality reduction and trimmed clustering. Random projection ensures computational tractability ($O(ndp)$) while preserving relative distances, and trimmed clustering isolates genuine structure from outliers.

The method's interpretability is an important advantage: HD-BWDM's scale directly compares between-cluster separation and within-cluster compactness, producing values that can be compared across different datasets or preprocessing schemes. The monotonic relationship with projection dimension also provides a diagnostic indicator for sufficiency of embedding dimension—an insight valuable for practitioners in large-scale data analysis.

Nevertheless, limitations remain. HD-BWDM's effectiveness depends on an adequate initial clustering method; if clustering fails entirely, the index cannot recover the correct structure. Moreover, when contamination exceeds 20–25%, even trimmed medoid-based approaches degrade. Incorporating robust covariance estimators or subspace modelling could further enhance resilience.

## 6. Conclusion

This study introduced HD-BWDM, a high-dimensional extension of the Between–Within Distance Median index for robust cluster validation. By integrating random or principal component projections, trimmed clustering, and medoid-based distance measures, HD-BWDM retains theoretical consistency and practical robustness in high-dimensional and contaminated data.

Simulations up to 500 dimensions demonstrated that HD-BWDM grows monotonically with projection dimension, effectively differentiates cluster quality, and remains stable under outlier contamination. These properties make HD-BWDM a valuable, computationally efficient stopping rule for modern high-dimensional clustering applications, including bioinformatics, text mining, and image analytics.

Future work will extend HD-BWDM to streaming and temporal data, where projections and medoid updates can be performed online.